\documentclass{article}
\usepackage{amsmath,epsfig}
\usepackage[preprint]{spconfa4}

\let\OLDthebibliography\thebibliography
\renewcommand\thebibliography[1]{
  \OLDthebibliography{#1}
  \setlength{\parskip}{0pt}
  \setlength{\itemsep}{0pt plus 0.3ex}
}

\begin{document}\sloppy

\def\x{{\mathbf x}}
\def\L{{\cal L}}

\title{PS-RCNN: Detecting Secondary Human Instances in a Crowd via Primary Object Suppression}
%
\name{Zheng Ge$^{\ast}$, Zequn Jie$^{\dagger}$, Xin Huang$^{\ast}$, Rong Xu$^{\ast}$ and Osamu Yoshie$^{\ast}$}

\address{$^\ast$Waseda University, $^\dagger$National University of Singapore \\
         \textit{\{jokerzz@fuji, koushin@toki, yoshie@\}.waseda.jp, \{zequn.nus,  xurong1981\}@gmail.com}}

\maketitle

\begin{abstract}
Detecting human bodies in highly crowded scenes is a challenging problem. Two main reasons result in such a problem: 1). weak visual cues of heavily occluded instances can hardly provide sufficient information for accurate detection; 2). heavily occluded instances are easier to be suppressed by Non-Maximum-Suppression (NMS). To address these two issues, we introduce a variant of two-stage detectors called PS-RCNN. PS-RCNN first detects slightly/none occluded objects by an R-CNN~\cite{girshick2014rich} module (referred as P-RCNN), and then suppress the detected instances by human-shaped masks so that the features of heavily occluded instances can stand out. After that, PS-RCNN utilizes another R-CNN module specialized in heavily occluded human detection (referred as S-RCNN) to detect the rest missed objects by P-RCNN. Final results are the ensemble of the outputs from these two R-CNNs. Moreover, we introduce a \textbf{H}igh \textbf{R}esolution \textbf{R}oI \textbf{A}lign (\textbf{HRRA}) module to retain as much of fine-grained features of visible parts of the heavily occluded humans as possible. Our PS-RCNN significantly improves recall and AP by 4.49\% and 2.92\% respectively on CrowdHuman~\cite{CrowdHuman}, compared to the baseline. Similar improvements on Widerperson~\cite{Zhang2019WiderPerson} are also achieved by the PS-RCNN. 
\end{abstract}
\section{Introduction}
\label{sec:intro}
\vspace{-0.2cm}

Human body detection is one of the most important research fields in computer vision. Although rapid developments have been seen in recent years, detecting human in crowded scenarios with various gestures still remains challenging. In the crowded situation, occlusions between human instances are common, making the visual patterns of occluded humans less discriminative and hard to be detected. In this work, we call those slightly/none occluded human instances as ``Primary Objects'', and those heavily occluded human instances as ``Secondary Objects'', and abbreviate them to P-Objects and S-objects, respectively. 

\begin{figure}[!t]
\begin{center}
\includegraphics[width=8.4cm,height=7cm]{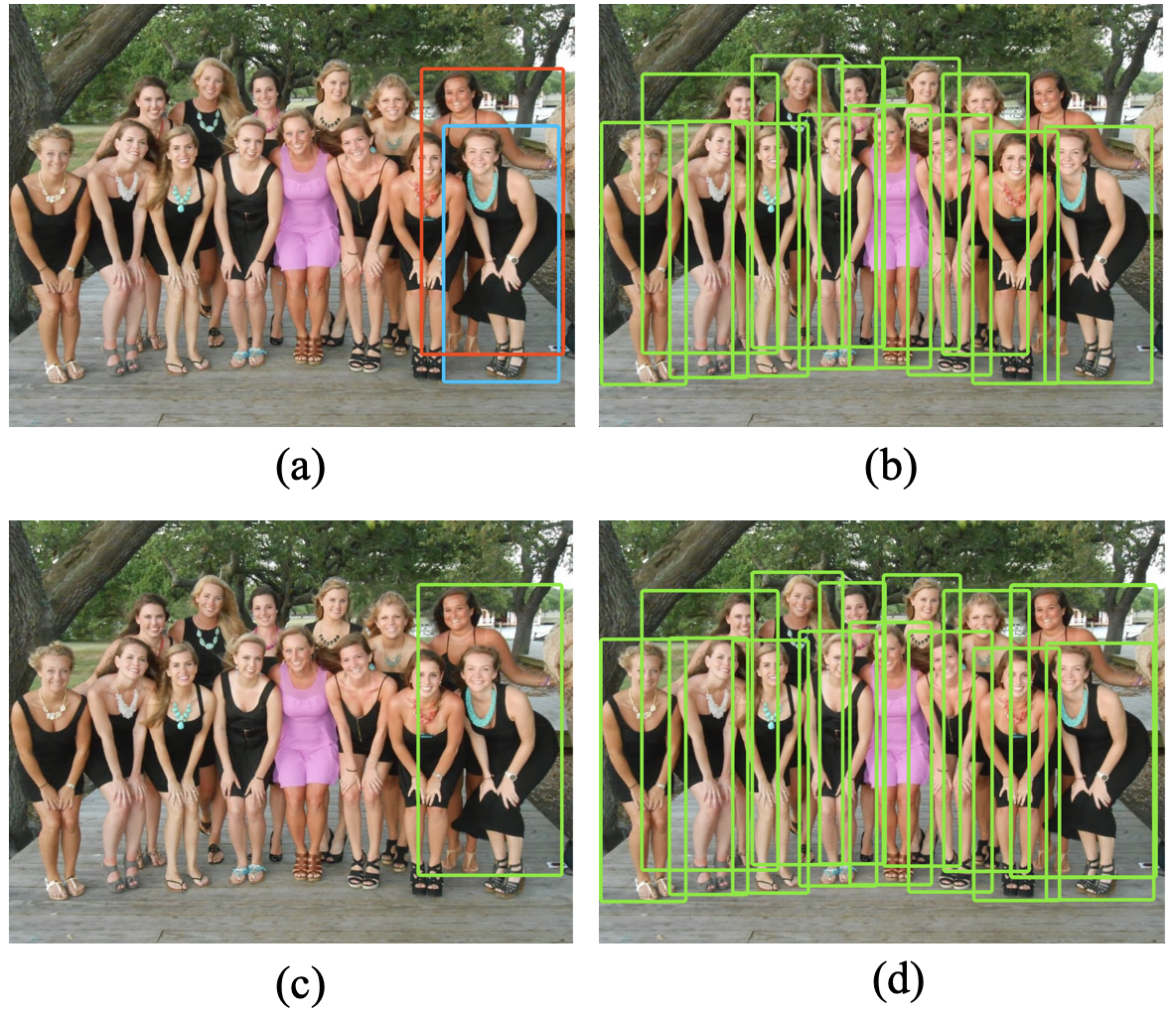}
\end{center}
   \caption{(a). The BBox in red will be suppressed by the BBox in blue if NMS@0.5 is applied. (b). Illustration of detected instances from P-RCNN. (c). Illustration of detected instances from S-RCNN. (d). Final outputs of PS-RCNN.}
   \label{fstart}
\end{figure}

There are two reasons leading to the poor detection performance on S-Objects. First, visual cues for S-Objects are weak compared to P-Objects which make them hard to be distinguished from backgrounds or other human bodies. Besides, the remarkable difference of visual features between P-Objects and S-Objects poses great challenge for a single detector in handling both of them. A preliminary experiment verifies such an observation. We evaluate on both P-Objects and S-Objects with a Faster R-CNN~\cite{Ren2017Faster} trained on CrowdHuman~\cite{CrowdHuman}. Results show that the missing rate for P-Objects and S-Objects are 4\% and 18\%, respectively, which reveals the poor detection performance on S-Objects. Second, S-Objects usually share large overlaps with other human instances, and thus they are easier to be treated as duplicate detections and then get suppressed during post-processing (\emph{i.e.} NMS). To verify this point, we print the statistics of pair-wise IoU in CrowdHuman and WiderPerson~\cite{Zhang2019WiderPerson} in Table \ref{pairwiseoverlaps}. As shown in Table \ref{pairwiseoverlaps}, there are 2.4 pairs of human instances with their overlaps (IoU) over 0.5 in CrowdHuman dataset. It means if we use NMS with the IoU threshold of 0.5 (NMS@0.5) as the default post-processing method, at least 2.4 instances will be inevitably missed per-image. Such missing detections can never be made up by strengthening the detector, while using a high IoU threshold like 0.7 will introduce a great number of false positives in the final results. Similar phenomenon can be observed on the WiderPerson.

To deal with the first issue, OR-CNN~\cite{Zhang2018Occlusion} proposes a new aggregation loss and occlusion-aware RoI Pooling to encourage the model to learn the existence of human bodies based on the occurrence of different human body parts. RepLoss~\cite{Repulsionloss} imposes additional penalty terms on the BBoxes that appear in the middle of two persons to achieve more accurate localization accuracy in crowded scenes. However, these methods do not pay any attention to the post-processing methods, which makes their models suffer from a hidden upper boundary of recall rate.

To alleviate the second issue, instead of discarding the suppressed BBoxes, Soft-NMS~\cite{softnms} lowers the classification scores for overlapped BBoxes according to their overlaps with the most confident one. Adaptive NMS~\cite{Liu2019Adaptive} argues that a uniform threshold of NMS is not suitable in crowded scenes. They predict a density map to allow NMS to run with different thresholds at different locations according to the spatial density of ground-truth objects. While these methods can improve the recall of detectors, they also suffer from a high risk of introducing lots of false positives, which makes them become sub-optimal solutions.

In this paper, we propose PS-RCNN, a variant of two-stage detectors to address the above two issues. PS-RCNN consists of two R-CNN modules on top of a shared backbone. The first R-CNN module which is trained in the exact same way with the standard Faster R-CNN is called Primary R-CNN Module (\emph{i.e.} P-RCNN). It aims to detect the primary human instances. The detected primary instances are then suppressed to facilitate the overlapped secondary human detection. To this end, we introduce a primary instance binary mask which effectively erases the primary instance, such that the weak feature of the occluded secondary instance can stand out. The second R-CNN module (\emph{i.e.} S-RCNN) which is specialized in occluded human detection, is then introduced to detect the secondary instances, based on the modified features. The whole structure of PS-RCNN can been seen in Fig.~\ref{fstart}. PS-RCNN can be trained in an end-to-end manner. By avoiding NMS@0.5 being the last post-processing method, our PS-RCNN can yield denser predictions. Moreover, since each R-CNN module is only responsible for detecting one kind of human instances (slightly/none or heavily occluded instances), the individual task of both primary object detection and secondary object detection can be improved.

\begin{table}[t]
\centering
\begin{tabular}{c|cc}
pair/image& Crowdhuman& Widerperson\\ \hline
IoU\textgreater 0.3 & 9.02& 9.21\\
IoU\textgreater 0.4& 4.89& 4.78\\
IoU\textgreater 0.5& 2.40& 2.15\\
IoU\textgreater 0.6& 1.01& 0.81\\
\end{tabular}
\caption{Pair-wise overlaps between human instances in CrowdHuman and WiderPerson datasets.}
\label{pairwiseoverlaps}
\end{table}

Our baseline is the naive Faster R-CNN with FPN~\cite{Lin2016Feature}, which augments a standard CNN with a top-down pathway, each level of which can be used for detecting objects at different scales. To further improve the visibility of S-Objects, we introduce High Resolution RoI Align (HRRA) module which only extracts features from the layer with the highest resolution. We also incorporate COCOPerson~\cite{lin2014microsoft} into our training pipeline to achieve more accurate instance masks. Combining the used of HRRA and training with COCOPerson leads to an AP improvement of 2.41\% on CrowdHuman.

To summarize, our contributions are as follows: (1) a novel PS-RCNN to fight against poor performance of S-Objects in crowded human detection; (2) a High Resolution RoI Align (HRRA) module to improve the visibility of S-Objects to detectors; (3) state-of-the-art performance of recall and great improvements of AP on both CrowdHuman and WiderPerson datasets.

\section{Related Work}
\vspace{-0.2cm}

\noindent \textbf{General object detection}. As one of the predominant detectors, Faster R-CNN~\cite{Ren2017Faster} first generates a set of region proposals and then refines them by a post classification and regression network. FPN \cite{Dollar2014Fast} extends Faster R-CNN by introducing a top-down pathway of features to tackle large scale variance of target objects. Mask R-CNN \cite{He2017Mask} proposes RoI Align and uses another instance segmentation branch to improve the localization ability of detectors. Cascade R-CNN~\cite{cai2018cascade} proposes a multi-stage R-CNN module trained with a set of increasing IoU thresholds, which is able to refine predict bounding boxes stage-wise.

\begin{figure*}[!t]
\begin{center}
\includegraphics[width=0.97\linewidth]{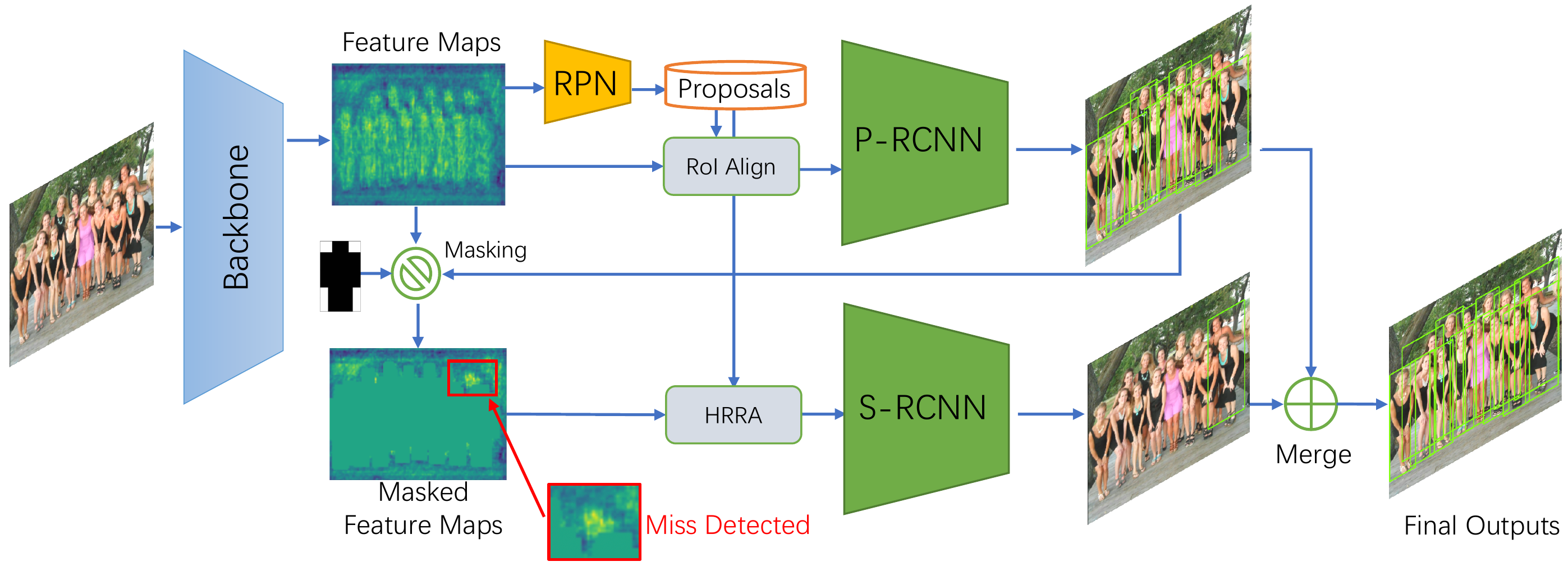}
\end{center}
 \vspace{-0.3cm}
   \caption{The framework of PS-RCNN. }
   \label{fstructure}
\vspace{-0.5cm}
\end{figure*}

\noindent \textbf{Occlusion handling}. Crowd occlusion (intra-class occlusion and inter-class occlusion) is the main challenge in pedestrian detection. 
\cite{CrowdHuman} and \cite{Zhang2019WiderPerson} propose two crowd human datasets (\emph{i.e.} CrowdHuman and WiderPerson) to better evaluate detectors in crowded scenarios. \cite{8578829} adopts attention mechanism across channels to represent various occlusion patterns. In \cite{6248062,8237639}, part-based model is utilized to handle occluded pedestrians. \cite{li2018multispectral} starts from the condition with insufficient illumination, then generates pedestrian proposals by a multi-spectral proposal network and proposes a subsequent multi spectral classification network. Occlusion-aware R-CNN\cite{Zhang2018Occlusion} designs an aggregation loss and an Occlusion-aware RoI Pooling to help detectors notice the existence of human instances based on the occurrence of human body parts. RepLoss \cite{Repulsionloss} proposes additional penalties to the BBoxes which appear in the middle of two persons to force predicted BBoxes locate firmly and compactly to the ground-truth objects. Adaptive NMS \cite{Liu2019Adaptive} predicts a density map of an image to perform a modified version of NMS with dynamic suppression thresholds. \cite{pang2019maskguided} incorporates mask guided attention to help model focus on visible regions of proposals to obtain better classification prediction.

\section{PS-RCNN}
\vspace{-0.2cm}

We propose PS-RCNN, a variant of two-stage detectors to handle the human body detection task in crowded scenarios. We further propose a High Resolution RoI Align (HRRA) module to help Secondary R-CNN Mudole (S-RCNN) read S-Objects at a layer with higher resolution. Finally, we introduce COCOPerson into our training pipeline to improve the quality of human-shaped masks, which leads to better performance of S-RCNN.

\subsection{Structure of PS-RCNN}

The structure of PS-RCNN can be seen in Fig.~\ref{fstructure}, PS-RCNN contains two parallel R-CNN modules (\emph{i.e.} P-RCNN and S-RCNN) on a shared backbone. Inspired by the merit of divide-and-conquer, P-RCNN is designed to detect those slightly/none occluded human instances (P-Objects) while S-RCNN needs to pick up those missed instances which are heavily occluded (S-Objects). 

In the naive Faster R-CNN, RPN~\cite{Ren2017Faster} needs to generate proposals with respect to all the ground-truth targets. In PS-RCNN, RPN plays the same role with Faster R-CNN, aiming to provide proposals w.r.t. all the human instances. The two R-CNN modules in PS-RCNN are designed to detect two sets of human instances with different occlusions. P-RCNN carries out the first round of detection, aiming to detect non/slightly occluded instances. S-RCNN is then leveraged to detect the heavily occluded ones missed by P-RCNN. Therefore, we train P-RCNN with all the ground-truth instances, while train S-RCNN using only the missed ones by P-RCNN to make it more complementary with P-RCNN. Formally speaking, we define $G = \{bbox_1, bbox_2, ..., bbox_n\}$ as all the ground-truth bounding boxes for image $I$. In the training phase, $G$ is used as the ground-truth objects for P-RCNN. The instances detected by P-RCNN are defined as $G_d$, and the missed instances are defined as $G_m$. After $G_d$ is detected, a human-shaped binary mask is covered at the location of ground-truth of each detected BBox in $G_d$ on feature maps, which can be seen in Fig.~\ref{fstructure}. After binary masking, the visual cues for $G_d$ should be invisible for S-RCNN. Thus only the missed instances $G_m$ are set as ground-truth objects to train S-RCNN. In this way, the two R-CNNs handle two separate sets of human instances, following in a divide-and-conquer manner. What's more, the division of non-occluded and occluded instances are automatically defined by the model itself, instead of using a hand-designed hard occlusion thresholding strategy. The final outputs of PS-RCNN are the union of the results from two R-CNN modules.

\subsection{High resolution RoI Align}

Feature Pyramid Network (FPN~\cite{Lin2016Feature}) is an effective way to handle large scale variance of objects in the real world, thus it has become a standard module in many recent state-of-the-art detectors. We also incorporate FPN into our PS-RCNN. FPN assigns objects with different scales to different levels of feature maps. Specifically, the larger object's scale is, the lower resolution level it will be assigned to for richer semantic information. Small objects will be assigned to feature maps at layers in higher resolution level because they may become indistinguishable if not so. An RoI Extractor (\emph{e.g.} RoI Align~\cite{He2017Mask}) is then used to extract features of these RoIs from different feature levels according to their scales. In our PS-RCNN, the RoI extractor for P-RCNN also follows this standard routine. However, when extracting RoI features for S-RCNN, we only extract features from the feature maps in the highest resolution level. Such an operation is called High Resolution RoI Align (HRRA). The introduction of HRRA module is based on an observation that though the full body of S-Objects are usually large, their informative visible regions could be extreme small. Directly assigning S-Objects to the layers in small resolutions according to the full-body scales would make the informative visible regions become even smaller and hard to identify. Therefore, we propose to perform RoI extraction of S-Objects from feature maps in higher resolution.

\begin{table*}[!t]
\centering
\caption{
Experimental results on CrowdHuman \textit{val set}. `*' denotes our re-implemented results.
Two region feature extraction methods in the rows of PS-RCNN stand for the RoI feature extraction method for P-RCNN and S-RCNN, respectively.}
\begin{tabular}{l|c|cc|cc}
\hline
Method         & Region Feature Extraction & AP    & Recall & $\Delta$ AP & $\Delta$ Recall \\ \hline
Baseline \emph{Shao et al.~\cite{CrowdHuman}}         &           RoI Pooling~\cite{Ren2017Faster}  &   84.95    & 90.24       &  -0.07  &  -0.38      \\
Baseline*      &           RoI Align               &    85.02   &  90.62      &  -  & -       \\
Adaptive NMS~\cite{Liu2019Adaptive}   &            -               & 84.71 & 91.27  &  -0.31   & 1.03    \\
Repulsion Loss*~\cite{Repulsionloss}  &           RoI Align               &   85.71    &  90.74      & 0.69   & 0.12       \\
Soft-NMS*~\cite{softnms}  &           RoI Align               &   85.66    &  92.88      & 0.64   & 2.26       \\ \hline
PS-RCNN        &           RoI Align; RoI Align                &  85.53     &  92.67      & 0.51   & 2.05       \\
PS-RCNN        &           RoI Align; HRRA          &  86.05     &  93.77      & 1.03   & 3.15       \\
PS-RCNN \textit{w/} IM   &           RoI Align; HRRA               &   \textbf{87.94}    &   \textbf{95.11}   &  \textbf{2.92}   & \textbf{4.49}      
\end{tabular}\label{mainresults}
\vspace{-0.4cm}
\end{table*}

\subsection{Higher quality instance masks from COCOPerson}

The quality of human-shaped binary masks is crucial. An imperfect mask which leaves parts of a primary instance uncovered will lead to plenty of duplicate detection results in the S-RCNN detection stage, bringing about lots of false positives. A hand-crafted human-shaped mask is qualified in most cases. However, it can not handle complicated cases when human body performs some uncommon gestures like dancing or bending their arms on others' shoulders, etc. Incorporating instance segmentation to acquire more accurate instance masks can alleviate this issue to some extent. Thus beyond the hand-crafted binary masks, we propose another enhanced version of PS-RCNN, which incorporates an instance segmentation branch after P-RCNN module to get instance masks for P-Objects. The instance segmentation branch is trained on COCOPerson and can be easily plugged into current PS-RCNN. S-RCNN does not need to be finetuned after the introduction of instance branch according to our experiments.

\section{Experiments}
\vspace{-0.2cm}

To validate the effectiveness of our proposed PS-RCNN, we conduct experiments on two crowded human detection datasets: CrowdHuman~\cite{CrowdHuman} and WiderPerson~\cite{Zhang2019WiderPerson}.

\subsection{Evaluation metrics}

The log average missing rate over 9 points ranging from 10$^{-2}$ to 10$^0$ FPPI (\emph{i.e.} mMR~\cite{dollar2011pedestrian}) is broadly used in pedestrian detection field. The score threshold for mMR is dynamic and usually determined by the number of false positive predictions. mMR is qualified in common pedestrian detection scenarios because it achieves a good balance between the number of true positive and false positive predictions. However, when detecting human body in crowded scenarios with various gestures, detectors may generate more false positives because the heavy overlaps between human instances. Thus mMR usually yields extremely high score thresholds when dealing with crowded scenes. Specifically, in our experiments, we find that the score threshold for mMR can reach 0.95 which means the predicted BBoxes whose class scores are less than 0.95 will not be considered in the stage of evaluation. Such a high score threshold is not able to fairly reflect the capability of human body detectors. Thus in this work, we use COCO~\cite{lin2014microsoft} style AP at IoU threshold of 0.5 and recall as our evaluation metrics. 

\subsection{Experiments on CrowdHuman}

\noindent \textbf{Dataset.} As a benchmark for detecting human body in the crowded situation, CrowdHuman provides three kinds of annotations: \textit{full body}, \textit{visible body} and \textit{head} to help researchers fully explore the potential of human body detection. It contains $15,000$, $4,370$, and $5,000$ images for training, validation and testing, respectively. On average, there are around 23 persons per-image, making CrowdHuman a challenging benchmark. We conduct all of our experiments on \textit{full body}, because to acquire the BBox for \textit{full body}, the detector not only needs to localize where the visible regions are, but also needs to estimate the boundaries of occluded areas, which makes it more challenging.

\noindent \textbf{Implementation details.} We adopt FPN with ResNet-50~\cite{he2016deep} as our baseline. RoI-Align is also used for better RoI feature alignment. For all of our experiments, we train detectors on 8 GPUs (2 images per-GPU) with an initial learning rate of 0.02, and decrease it by 0.1 at the 8th and 11th epoch. The training process finishes at the end of the 12th epoch. Standard SGD with momentum 0.9 is used as the optimizer. We use the same set of anchor ratios with \cite{CrowdHuman} which are {1.0,1.5,2.0,2.5,3.0}. The input image is re-scaled such that its shortest edge is 800 pixels, and the longest side is not beyond 1400 pixels, considering the large scale variance of images in CrowdHuman. For the enhanced version of PS-RCNN, we train the additional instance segmentation branch with extra 9 epochs on COCOPerson, during which only the parameters of instance prediction branch are updated. Since the R-CNN module we use in our model only contains a few fully-connected layers, so our PS-RCNN only brings limited extra computation cost.

\noindent \textbf{Main results.} Results on CrowdHuman are presented in Table \ref{mainresults}. Our re-implemented baseline is slightly better than \cite{CrowdHuman}, in which we replace RoI Pooling with RoI Align. ``IM'' is the abbreviation of ``instance masks''. ``Soft-NMS'' is implemented by simply replacing NMS in ``Baseline'' with ``Soft-NMS''. As can be seen in Table \ref{mainresults}, without HRRA for S-RCNN, the improvement on recall is only 2.05\% because the visual cues of S-Objects are very weak. After adopting HRRA, our PS-RCNN with HRRA can bring 1.03\% and 3.15\% improvements on AP and recall, respectively. Moreover, after replacing the hand-crafted human-shaped masks with predicted instance masks, total improvements on AP and recall reach 2.92\% and 4.49\%. It is worth mentioning that if we feed ground-truth annotations of \textit{full body} into NMS@0.5, the recall of the final outputs is 90.9\%\footnote{Slightly different values of recall are possible because we set scores for all ground-truth BBoxes as 1.0, so that the order of BBoxes is random.} which means using NMS@0.5 as the final post-processing method can hardly surpass the recall of 90.9\%. Soft-NMS and Adaptive NMS break this hidden upper limit of recall to some extent. Our method further pushes the state-of-the-art of recall to 95.11\%.

\begin{figure}[h]
\begin{center}
\includegraphics[width=1.0\linewidth]{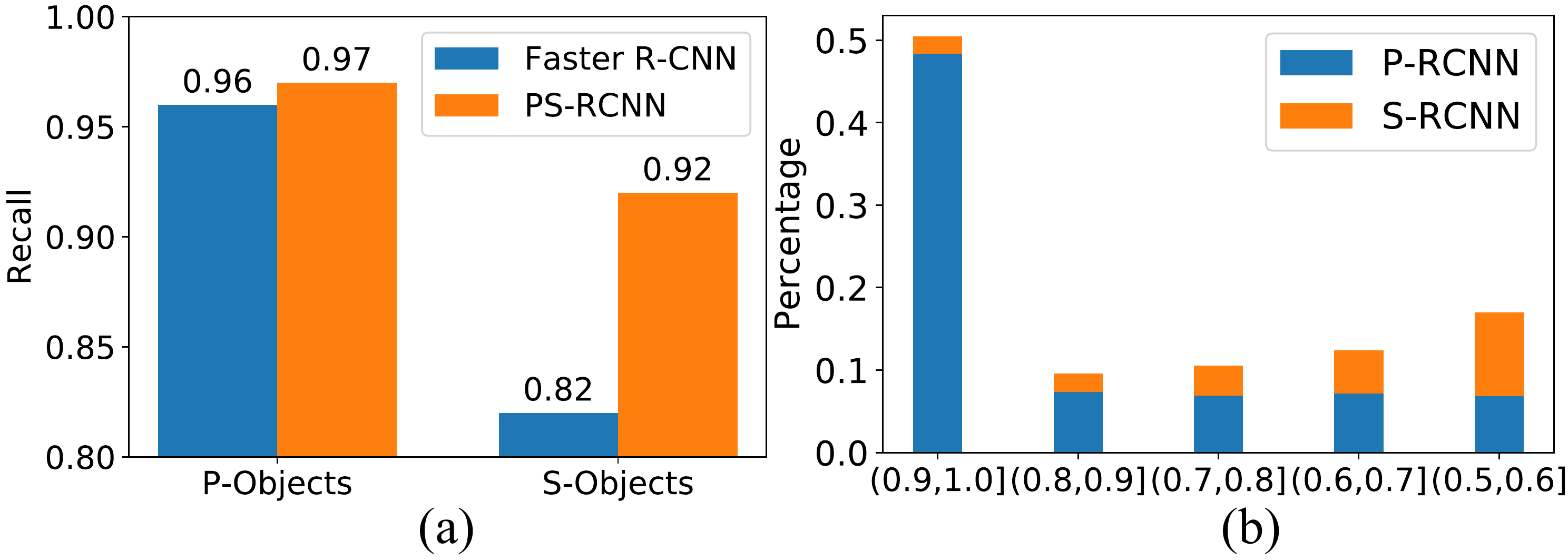}
\end{center}
\vspace{-0.3cm}
   \caption{(a). Recall of P-Objects and S-Objects for Faster R-CNN and PS-RCNN. (b). Distribution of predicted scores from P-RCNN and S-RCNN. Only predicted BBoxes whose score$>$0.5 are counted.}
   \label{fdistinguish}
   \vspace{-0.3cm}
\end{figure}

\noindent \textbf{Discussion.} To validate the effectiveness of our PS-RCNN on detecting S-Objects, we count the number of detected P-Objects and S-Objects before and after adopting PS-RCNN. Because there is no clear definition on P-Objects and S-Objects, here we define a term $V$ to represent the degree of visibility, where $V={S_v}/{S_f}$. $S_v$ and $S_f$ stand for the area of visible body region and full body region, respectively. As can be seen in Fig.~\ref{fdistinguish}(a), PS-RCNN improves the recall of instances whose $V<=0.5$ by 10\% (0.82 \emph{v.s.} 0.92) compared to its baseline, while the recall for instances whose $V>0.5$ also increases a little. Moreover, to prove that our PS-RCNN does not work like Soft-NMS which keeps plenty of detected BBoxes with extremely low confidence scores, we plot the distribution of confidence scores from both P-RCNN and S-RCNN in Fig.~\ref{fdistinguish}(b). From Fig.~\ref{fdistinguish}(b), we can draw a conclusion that S-RCNN actually yields considerable amount of high score predictions which can be beneficial to real applications.

\begin{table}[]
\caption{Ablation study about where to apply human-shaped masks on.}
\centering
\begin{tabular}{c|cc}
\hline
humanoid masks  & AP & Recall \\ \hline
on input image & 87.86   & \textbf{95.40}       \\
on feature maps   & \textbf{87.94}   & 95.11       \\ \hline
\end{tabular}\label{fori}
\vspace{-0.3cm}
\end{table}

\noindent \textbf{Ablation study.} The effectiveness of high quality instance masks and HRRA has been verified in Table \ref{mainresults}. Here, we explore other two factors that might influence the performance of PS-RCNN. The first is where should instance masks be applied -- original image or feature maps? Table \ref{fori} shows that applying masks on original image or feature maps does not make much difference. But considering time efficiency, directly applying masks on feature maps is preferred in real applications. The second is which IoU threshold should we use to define positive proposals for S-RCNN? S-Objects are usually heavily occluded. Estimating the BBoxes for them are tougher than P-Objects, thus better proposals may help. We try 0.5, 0.6 and 0.7 as the IoU thresholds of positive proposals for S-RCNN. Results in Table \ref{thre} show that IoU threshold of 0.5 yields better recall while 0.6 achieves the best AP. AP is usually more important than Recall, so we recommend 0.6 as IoU threshold of positive proposals when using PS-RCNN.

\begin{table}[]
\caption{Varying the IoU threshold for positive proposals when training S-RCNN.}
\centering
\begin{tabular}{c|cc}
\hline
threshold & AP    & Recall \\ \hline
0.5       & 87.53 & \textbf{95.46}  \\
0.6       & \textbf{87.94} & 95.11  \\
0.7       & 86.91 & 93.50  \\ \hline
\end{tabular}\label{thre}
\vspace{-0.5cm}
\end{table}

\vspace{-0.3cm}
\subsection{Experiments on WiderPerson}
\vspace{-0.2cm}
WiderPerson~\cite{Zhang2019WiderPerson} is another dense human detection dataset which is collected from various kinds of scenarios. It contains five types of annotations -- pedestrians, riders, partially-visible persons, crowd and ignored regions. In our experiments, we merge the former four types into one category in both training and testing phase. WiderPerson contains 8,000, 1,000 and 4,382 images for training, validation and testing sets. We use the exactly same division as the original widerperson dataset. We train our PS-RCNN on the training set and validate on the validation set. The other settings follow our experiments on CrowdHuman. Results are presented in Table \ref{widerperson}. As seen in Table \ref{widerperson}, our PS-RCNN with HRRA and IM improves AP and recall by 1.63\% and 2.01\%, respectively, verifying the effectiveness of our proposed PS-RCNN across different datasets.

\begin{table}[!h]
\caption{Experimental results on WiderPerson.}
\centering
\begin{tabular}{l|cc}
\hline
Method                 & AP    & Recall \\ \hline
Faster R-CNN (Ours)    & 88.89 & 93.60  \\ \hline
PS-RCNN                & 89.34 & 94.36  \\
PS-RCNN w/ HRRA        & 89.96 & 94.71  \\
PS-RCNN w/ HRRA and IM & \textbf{90.52} & \textbf{95.61}  \\ \hline
\end{tabular}\label{widerperson}
\vspace{-0.4cm}
\end{table}

\section{Conclusions}

In this paper, we propose a new two-stage human body detector called PS-RCNN, which utilizes two parallel R-CNN modules to detect slightly/none occluded and heavily occluded human instances. We also introduce high quality instance masks into our model via few extra epochs of training on COCOPerson and a High Resolution RoI Align (HRRA) module to improve the capability of Secondary R-CNN module. Experiments on CrowdHuman show that our PS-RCNN can totally improve the AP and recall by 2.92\% and 4.49\%, respectively. Great improvements can also be observed on WiderPerson dataset, showing the effectiveness of PS-RCNN.

\bibliographystyle{IEEEbib}
\bibliography{icme2020template}

\end{document}